\documentclass{article}
\usepackage{ijcai19}
\usepackage{times}
\usepackage{soul}
\usepackage{url}
\usepackage[utf8]{inputenc}
\usepackage[small]{caption}
\usepackage{graphicx}
\usepackage{amsmath}
\usepackage{booktabs}
\usepackage{subfig}
\usepackage{ijcai19}
\usepackage{times}
\usepackage{epsfig}
\usepackage{graphicx}
\usepackage{float}
\usepackage{amsmath}
\usepackage{amssymb}
\usepackage{algorithm}
\usepackage{algorithmic}
\usepackage{subfig}
\usepackage{slashbox}
\usepackage{booktabs}
\usepackage{multirow}
\usepackage{enumerate}
\usepackage{array}
\usepackage{bm}
\renewcommand{\paragraph}{\textbf}

\title{Extracting Visual Knowledge from the Internet: Making Sense of Image Data}

\author{
Yazhou Yao$^{12}$\and
Jian Zhang$^{1}$\and
Xiansheng Hua$^{3}$\and
Fumin Shen$^{4}$\and
Zhenmin Tang$^{2}$\\
\affiliations
$^1$University of Technology Sydney
$^2$Nanjing University of Science and Technology\\
$^3$Alibaba Group
$^4$University of Electronic Science and Technology of China\\
}
\begin{document}

\maketitle

\begin{abstract}
Recent successes in visual recognition can be primarily attributed to feature representation, learning algorithms, and the ever-increasing size of labeled training data. Extensive research has been devoted to the first two, but much less attention has been paid to the third. Due to the high cost of manual data labeling, the size of recent efforts such as ImageNet is still relatively small in respect to daily applications. In this work, we mainly focus on how to automatically generate identifying image data for a given visual concept on a vast scale. With the generated image data, we can train a robust recognition model for the given concept. We evaluate the proposed webly supervised approach on the benchmark Pascal VOC 2007 dataset and the results demonstrates the superiority of our method over many other state-of-the-art methods in image data collection.
\end{abstract}

\section{Introduction}

With the development of the Internet, we have entered the era of big data. It is consequently a natural idea to leverage the large scale yet noisy data on the Internet \cite{yao2018aaai,shen2018tmm,shen2018tmm2,fergus2010learning,fergus2004visual}. Methods of utilizing these data for visual recognition have recently become a hot topic, with the convergence of computer vision, pattern recognition and machine learning being collectively known as ``Internet vision'' \cite{fergus2010learning}.

When generating massive identifying image data, it is important to ensure that the data contains sufficient representative visual patterns. Search engines (e.g., Google Images) and social media portals (e.g., Flickr) have been created to obtain the candidate image data \cite{xuicme2016,liutip2019,tangijcai2018,michel2011quantitative,schroff2011harvesting}. For example, we can submit a certain concept as a query to Google Images or Flickr which will return a large number of images based on that query. However, the images returned by search engines or social media portals are usually attractive and representative, but they incorporate fewer visual patterns. For example, if we submit the term ``tiger'' to Google Images, most of the returned images are of tiger faces. To build a high-quality training dataset, however, we need to collect a large number of Internet images which contain different views of an image as visual patterns. To address this problem, we propose a two-step approach in this paper, as follows: We first find the useful related word variations to enrich the given concept from a text perspective, which is known as label purification; we then use these selected word variations to collect images and run a further image clean process, which is known as image purification. Our goal is to achieve a certain level of match between the image labels and their dominant contents to narrow the semantic gap.

Search results returned from Google Books Ngrams corpora are usually very noisy \cite{huatmm2017,zhang2016domain,prl2018}. For example, word variations returned from Google Books Ngrams corpora for the query ``horse'' contains not only different``visual patterns'' of a horse such as a ``jumping horse'' and a ``rearing horse'', but also a large number of different word variations associated with ``horse'' such as ``horse boy'' and ``horse stealer''. Our starting point in this step is to find key``visual patterns''of horse. Therefore, ``horse boy'', ``horse stealer'' etc should be removed from the word variations list.

In addition, images returned from the Google search engine also contain noise. For example, it may contain some clipart images even at the top of the return list. In order to build a large scale robust image dataset, noisy images such as clipart images should be filtered out without too much manual effort.  

By searching in the Google Books Ngrams corpora and Google images, it is easy to get over 1000 related word variations and 10000 images of the given concept. However, how to effectively leverage this massive and noisy data to build a robust image dataset remains a huge challenge. In this paper, we argue that combining word variation purifying and image purifying is a more effective way to use this massive amount of data. The main contributions are: 1) Our method is the first to combine related word variation purifying and image purifying to build the image dataset. 2) Our method is able to build the image data for any given visual concept. 3) Our method is capable of training a robust recognition model without manual intervention.

\section{Related Work}

Recent work has mainly focused on learning from large datasets for recognition and classification \cite{deng2009imagenet,li2010optimol,perona2010vision,Fumin_2015_ICCV,Shen_2015_CVPR,shen2013inductive}. To our knowledge, there are three principal methods of building the database: manual annotation, active learning, and using Internet data.

\begin{table*}[t]
	\renewcommand{\arraystretch}{1.1}
	\centering
	\caption{Results of purifying noisy word variations using NGD for ``tiger'' and ``horse''.}
	\begin{tabular}{c|c|c|c|c|c|c|c}
		\hline
		& \multicolumn{3}{|c|}{Found word variations}  &\multicolumn{4}{|c}{after NGD filtering}  \\
		\hline
		Concept & correct & noisy & precision & correct & noisy & precision & false pos  \\
		\hline
		Horse   & 132     & 460   & 22.3\%    & 124     & 237   & 34.3\%    & 8  \\
		
		Tiger   & 108     & 536   & 16.8\%    & 102     & 287   & 26.2\%    & 6  \\
		\hline 	    	    	    	    	    	    	     	       						
	\end{tabular}
	\label{tab1}
\end{table*}

\subsubsection{Manual Annotation}

Manual annotation has a high level of accuracy but is resource-intensive, so the scale of the dataset is relatively small (both the numbers of categories and images) in the early years. The method of building ImageNet is using manual annotations. It firstly download images from the search engines using different languages for the given concept and then label these images by the power of crowds \cite{deng2009imagenet}.

\subsubsection{Active Learning}

To reduce the cost of manual annotation, recent work has also focused on active learning (a special case of semi-supervised learning) \cite{prest2012learning,siddiquie2010beyond,siva2011weakly,vijayanarasimhan2014large}, which selects label requests. \cite{prest2012learning} introduces an approach for learning object detectors from real-world web videos. The limitation is these web videos should only contain objects of a target class. \cite{siddiquie2010beyond} build a contextual object recognition model by using active learning. It needs the user's answer to update the existing object recognition model. \cite{siva2011weakly} learn a detector for a specific object class using weakly supervised learning and need a initial annotation. Both manual annotation and active learning require pre-existing annotations, which results in one of the biggest limitations to building a large scale dataset. 

\subsubsection{Using Internet Data}

Recent methods propose the use of Internet data to build the training data for visual concept \cite{chen2013neil,divvala2014learning,fergus2004visual,li2010optimol,schroff2011harvesting,mta2018,huangpu,yaotkde2019,xu2017,tang2017}. The general method is to automatically collect images using search engines or social networks to build a training set, and then to re-rank these images using visual classifiers \cite{divvala2014learning,fergus2004visual} or some form of clustering in visual space \cite{schroff2011harvesting}. \cite{chen2013neil} extracts visual knowledge from the pool of visual data on the web, mainly focusing on finding labeled segments/boundaries and relationships between objects. \cite{li2010optimol} uses iterative methods to automatically collect object datasets from the web and to incrementally learn object category models. This approach has some uncertainty, since it depends on the seed images. In addition, using this method to collect images of objects may contain fewer visual patterns. However, the use of ``re-ranking'', ``clustering'' and the other methods mentioned above does not tackle the problem of insufficient visual patterns in these training data. To address this problem, we propose our method for building a high-quality training dataset in the next section.

\section{The Proposed Approach}

Due to the amount and complexity of Internet data, to build a high-quality training dataset, we must separate noisy data from useful data automatically. Specifically, we first use Google Books Ngram corpora to obtain all the related word variations modifying the given concept; second, we use Normalized Google distance (NGD), linear SVM and visual distance to prune these word variations; third, we download images according to the filtered word variations, use exemplar-LDA and progressive CNN to purify these noisy images; lastly, we put all the purified images based on the various word variations together and fine-tune a robust model for object recognition. The following subsections describe the details of our process.

\subsection{Discovering Word Variations for the Given Concept}

As mentioned above, we use Google Books Ngrams corpora to generate ``visual patterns'' on the Internet scale for the given concept \cite{lin2012syntactic}. These corpora cover almost all related word variations for any concept at the text level and are much more general and richer than WordNet or Wikipedia \cite{miller1995wordnet,volkel2006semantic}. We use Google Books Ngrams corpora to discover related word variations for the given concept with parts-of-speech (POS), specifically with NOUN, VERB, ADJECTIVE and ADVERB. Using Google Books Ngrams data helps us cover all variations of any concept the human race has ever written down in books \cite{divvala2014learning}. In addition, using the POS tag can help us to partially purify word variations.

\subsection{Purifying Noisy Word Variations}
In our experiments, we found that not all the retrieved word variations are relevant to our given visual concept, e.g., ``tiger sharks'', ``tiger belles'', etc. Downloading images with these noisy word variations of the concept are harmful for training our target model ``tiger''; therefore, we need to purify these noisy word variations and set up key word variations. We developed a fast NGD method to address this issue from the perspective of text semantics. 

\subsubsection{Purifying Based on NGD}
Words and phrases acquire meaning from the way they are used in society, from their relative semantics to other words and phrases. For computers, the equivalent of ``society''is ``database'', and the equivalent of ``use'' is ``a way to search the database'' \cite{collosal2005well}. Normalized Google distance constructs a method to automatically extract similarity distance from the World Wide Web (WWW) using Google page counts \cite{cilibrasi2007google}. For a search term \emph{x} and search term \emph{y} (just the name for an object rather than the object itself), Normalized Google distance is defined by (1):
\begin{equation}NGD(x,y)=\frac{\max\{\log f(x),\log f(y)\}-\log f(x,y)}{\log N-\min\{\log f(x),\log f(y)\}}\end{equation}
where \emph{f(x)} denotes the number of pages containing \emph{x}, \emph{f(x,y)} denotes the number of pages containing both \emph{x} and \emph{y} and \emph{N} is the total number of web pages searched by Google.

\begin{table*}[t]
	\centering
	\renewcommand{\arraystretch}{1.1}
	\caption{Some examples of the merged word variations}
	\begin{tabular}{c|c}
		\hline
		Concept: & Merged word variations:  \\
		\hline
		Horse & \{horse cabs vs horse carriages vs horse carts vs horse buses vs horse trams\} \\			                           
		\hline
		Dog   & \{horse cabs vs horse carriages vs horse carts vs horse buses vs horse trams\} \\	
		
		\hline			
		Bus & \{double bus vs london bus vs open bus\} \\			                           
		\hline			
	\end{tabular}
	\label{tab3}
\end{table*}

\begin{table*}[t]
	\centering
	\renewcommand{\arraystretch}{1.1}
	\caption{Classification results of the concepts ``tiger'' and ``horse''}
	\begin{tabular}{c|p{2cm}<{\centering}|c|c}
		\hline
		Concept: & \emph{$S_i$} & Precision rate: & Recall rate:  \\
		\hline
		\multirow{3}{*}{Horse} & 0.71  & 0.8978 & 0.9278 \\			                           
		& 0.68   & 0.9408 & 0.8833  \\
		& 0.66  & 0.9615 & 0.8332 \\			                           
		\hline
		\multirow{3}{*}{Tiger} & 0.71 & 0.9045 & 0.9344  \\
		& 0.68  & 0.9523 & 0.8932 \\			                           
		& 0.66 & 0.9731 & 0.8374  \\
		\hline							
	\end{tabular}
	\label{add-tab}
\end{table*}

We denote the distance of all word variations by a graph \emph{$G_g = \{N, D\}$} where each node represents a word variation and its edge represents the \emph{NGD} between the two nodes. We set the target concept as center (\emph{$d_{0}$}) and other word variations have a score (\emph{$d_{x}$}) which corresponds to the distance to the target concept, (\emph{$d_{xy}$}) represents the \emph{NGD} between two word variations \emph{x, y}, and is defined as (2):
\begin{equation}d_{xy}=\frac{NGD(x,y)+NGD(y,x)}{2}\end{equation}

By setting the threshold $d_{x}$ to a value (0.5) we can successfully remove most of the noisy word variations while keeping the vast majority of useful word variations. Then we set $d_{xy}$ to a value (0.1) to merge semantic synonyms. This step can be viewed as part of a cascade strategy that purifies irrelevant word variations, so we set the value of $d_{x}$ and $d_{xy}$ to ensure the vast majority of useful word variations are passed on to the next stage. Table \ref{tab1} presents the results of purifying noisy word variations using NGD for the concept ``tiger'' and ``horse''. By experiments, we found that there are still lots of visual non-salient word variations, e.g., ``tiger shooting'', ``social tiger'', etc. and we cannot purify these noisy word variations using \emph{NGD} alone. 

\subsubsection{Purifying Based on Linear SVM}
From the visual perspective viewpoint, we want to identify visual salient word variations and eliminate visual non-salient word variations in this step. The intuition is that visual salient word variations should exhibit predictable visual patterns that are accessible to classifiers \cite{divvala2014learning}. We use the image-classifier based purifying method.

For each filtered word variation, we directly download 100 images from the Google image search engine as positive images; then randomly split these images into a training set (75 images) and validation set (25 images) \emph{$I_i=\{I_i^t, I_i^v\}$}, we gather a random pool of negative images (50 images) and split them into a training set (25 images) and validation set (25 images) \emph{$\overline I=\{\overline I^t, \overline I^v\}$};  We then train a linear SVM \emph{$C_i$} with \emph{$I_i^t$} and \emph{$\overline I^t$} , using dense HOG features and then use \emph{$\{I_i^v,\overline I^v\}$} as validation images to calculate the classification results \emph{$S_i$} \cite{fan2008liblinear,lin2011large}. Table \ref{add-tab} shows the results of the experiments on the concepts ``tiger'' and ``horse''. \emph{$S_i$} denotes the percentage of correctly classified images. We declare an word variation \emph{i} to be visually salient if the classification results \emph{$S_i$} give a relatively high score (0.7) as we want to purify visual non-salient word variations as much as possible.  

\subsubsection{Purifying Based on Visual Distance}
By filtering the noisy word variations based on NGD and linear SVM, we obtain a relatively clean word variations set. We found some word variations share similar visual patterns, e.g., ``tiger cubs, small tiger, baby tiger, little tiger''. To avoid collecting too many visually similar images as training data, we group these word variations \cite{malisiewicz2008recognition}.  We represent the distance of all word variations or labels by a graph \emph{$G_v = \{V,E\}$} where each node represents a word variation and each edge represents the distance between two variations. Each node has a score \emph{$V_i$} which corresponds to the classifier \emph{$C_i$} on its validation data \emph{$\{I_i^v,\overline I^v\}$}. As previously mentioned, the edge weights \emph{$E_{ij}$} from the visual perspective correspond to the distance between two word variations (labels) i, j and are measured by the score of the jth word variation classifier \emph{$C_j$} on the ith word variation validation set \emph{$\{I_i^v,\overline I^v\}$}. We group these word variations based on their distances to satisfy (3): 
\emph{\begin{equation}E_{ij} + 0.1 \geqslant V_j\end{equation}}
Table \ref{tab3} shows some examples of word variations merged by our method:

By using the above procedures, we obtain relatively clean word variations to represent different ``visual patterns'' for the given concept. Algorithm \ref{alg1} shows the process of purifying these irrelevant word variations.

We use these filtered word variations as queries to download the top 120 images from Google image for each word variation. Then we put these images together as an initial image dataset for the given concept. Although we get the initial images from the Internet similar to \cite{chen2013neil,divvala2014learning,fergus2010learning,fergus2004visual,li2010optimol,mezuman2012learning,rubinstein2013unsupervised,schroff2011harvesting}, the difference is we firstly get lots of related word variations representing different ``visual patterns'' for the given concept. Downloaded images using these related word variations are much more ample than only using the given concept. This is also our advantage over other methods. Table 4 shows the relatively clean word variations found by our ``purifying word variations":

\begin{table*}[t]
\centering
\renewcommand{\arraystretch}{1.1}
\caption{Word variations found on Pascal VOC 2007}
\begin{tabular}{c|c|c|c|c|c|c|c|c|c|c}
\hline
concept: & bottle & train & cat & cow & dog & horse & sheep & plane & bus & car\\
\hline
vars:    & 128    & 47    & 174 & 136 & 145 & 116 & 123 & 135 & 59 & 125\\
\hline
concept: & bike   & boat  & sofa & bird & mbike & plant & table & chair & pers & tv\\
\hline
vars:    & 43     & 207   & 38   & 226 & 33 & 5 & 437 & 345 & 209 & 36\\		
\hline
\end{tabular}	
\end{table*}

\subsubsection{The Limitation of Our Method}

From our experiments, we found our method is not able to remove these noisy word variations thoroughly. Also, some positive word variations may be filtered out incorrectly. Filtering our word variations using the previous steps results in an average (for PASCAL Visual Object Classes) of 3.24\% noisy word variations and an average 2.84\% positive word variations being filtered out for the given concept. Using these word variations may result in noisy images to our initial image dataset for the given concept (type 1 noisy images). We found these word variations are a very small number respect to the correct word variations. These few noisy images caused by noisy word variations can be effectively filtered out by the next image purifying steps. 

There are other types of noisy images which result from correct word variations in our initial image dataset for the given concept. Although the Google image search engine ranks the returned images, some noisy images are still included (type 2 noisy images). The reason for this is that the Google image search engine is a text based image search engine. To build a high-quality image dataset, both of these two types of noisy images should be removed from the initial image dataset.

\begin{algorithm}[t]
	\caption{Word variations purifying algorithm}
	\label{alg1}
	\begin{algorithmic}[1]
		\REQUIRE ~~\\
		\emph{$X = \{x_0\}$}, a concept for image label\\		
		\STATE Discover word variations in Google Books Ngrams corpora with POS and get \emph{$X = \{x_0, x_1,x_2....x_n\}$}\\
		\STATE Calculate NGD \emph{$d_{i}$} between \emph{$x_0$} and \emph{$x_i$}, delete \emph{$x_i$} from X if \emph{$d_{i} \geqslant 0.5$} \\	
		\STATE Calculate NGD \emph{$d_{ij}$} between \emph{$x_i$} and \emph{$x_j$} (\emph{$i,j \geqslant 0$}), merge \emph{$x_i$} and \emph{$x_j$} if \emph{$d_{ij} \leqslant 0.1$}\\	
		\STATE Delete visually non-salient \emph{$x_i$} from \emph{X} if  \emph{$S_i \leqslant 0.7$}\\		
		\STATE Merge visually similar \emph{$x_i$}  and \emph{$x_j$} if \emph{$E_{ij} + 0.1 \geqslant V_j$}\\
		\ENSURE ~~\\		
		a relatively clean word variation \emph{X} for the given concept		
	\end{algorithmic}
\end{algorithm}

\subsection{Purifying Noisy Images}

Most current approaches handle these problems via clustering \cite{kankanhalli1996cluster,lucchi2012joint,wang2007cell}. Clustering can help assist with handling visual diversity and can reject outliers based on their distance from cluster centers. However, clustering presents a scalability issue for our problem. Since our images are sourced directly from the Internet and have no bounding boxes, every image creates millions of data points, the majority of which are outliers. Recent work has suggested that K-means is not scalable \cite{doersch2012makes}. Instead, we propose to use a two-step approach to purify these noisy images.

\subsubsection{Purifying Based on Exemplar-LDA}

To purify these type 2 noisy images caused by Google image search engine, we download a set of images (120) from the Google image search engine for each selected word variation such as ``fighting tiger''. The image set is used to train a detector using exemplar-LDA \cite{hariharan2012discriminative}, and these detectors are then used for dense detections on the same image set. We select the top 100 images with high scores from multiple detectors for the next step. This method assists us to prune those images which relate less well to the word variations. 

\subsubsection{Purifying Based on Progressive CNN}

To reduce the influence of type 1 noisy images, we use a purifying method similar to that proposed by \cite{you2015robust}. The difference is that we do not train a CNN from the beginning; instead, we directly fine-tune a CNN on filtered images with a trained model ``$bvlc\rule[-2pt]{0.15cm}{0.5pt}reference\rule[-2pt]{0.15cm}{0.5pt}caffenet$"
\cite{jia2014caffe}. We then use the probabilistic sampling algorithm to select the new training samples according to the prediction score of the fine-tuned model on the training data itself. The intuition is we want to keep images with distinct sentiment scores between the two classes with high probability, and remove images with similar sentiment scores for both classes with high probability. Let \emph{$S_i = (S_{i1},S_{i2})$} be the prediction sentiment scores for the two classes of instance \emph{i}. We choose to remove the training instance \emph{i} with probability \emph{$P_i$} given by (4):
\emph{\begin{equation}P_i = max(0,2-exp(|S_{i1}-S_{i2}|))\end{equation}}
The training instance will be kept in the training set if the predicted scores of one training instance are large enough. Otherwise, the smaller the difference between the predicted scores, the large the probability that this instance will be removed from the training set. Type 1 noisy images can be effectively filtered by this step. The reason for this is that the number of this type of noisy images is relatively small in the whole image dataset for the given concept. 
Algorithm \ref{alg2} shows the detailed process for purifying noisy images.

\begin{algorithm}[t]
	\caption{Image purifying algorithm}
	\begin{algorithmic}[1]
		\REQUIRE ~~\\
		Word variations \emph{$X = \{x_0,x_1,x_2...\}$} as image labels for the given concept\\		
		\STATE Download 120 images for each selected word variation \emph{$x_i$} in X\\
		\STATE Select top 100 images with exemplar-LDA for each \emph{$x_i$}\\
		\STATE Purify the remaining noisy images with progressive CNN\\
		\ENSURE ~~\\		
		a relatively clean image dataset for the given concept		
	\end{algorithmic}
\label{alg2}
\end{algorithm}

\begin{table*}[t]
	\centering
	\renewcommand{\arraystretch}{1.1}
	\caption{Results (Average Precision) on Pascal VOC 2007 (test) object detection}
	\begin{tabular}{c|c|c|c|c|c|c|c|c|c|c|c}
		\hline
		Method & Supervised & bottle & train & cat & cow & dog & horse & sheep & plane & bus & car\\
		\hline
		\cite{siva2011weakly} & weak & 0 & \textbf{34.2} & 7.1 & 9.3 & 1.5 & 29.4 & 0.4 & 13.4 & 31.2 & \textbf{43.9}\\
		\hline		
		\cite{divvala2014learning} & web & \textbf{9.2} & 23.5 & 8.4 & 17.5 & 12.9 & 30.6 & \textbf{18.8} & 14.0 & 35 & 35.9\\
		\hline			
		\textbf{Ours} & \textbf{web} & 8.7 & 22.5 & \textbf{10.3} & \textbf{18.7} & \textbf{13.6} & \textbf{32.7} & 13.7 & 13.8 & \textbf{36.5} & 35.6\\	
		\hline
		\cite{felzenszwalb2010object} & full & 26.6 & 45.2 & 22.5 & 24.3 & 12.6 & 56.5 & 20.9 & 33.2 & 52.0 & 53.7\\	
		\hline
		\hline
		Method & Supervised & bike & boat & sofa & bird & mbike & plant & table & chair & pers & tv\\
		\hline
		\cite{siva2011weakly} & weak & \textbf{44.2} & 3.1 & 3.8 & 3.1 & \textbf{38.3} & 0.1 & \textbf{9.9} & 0.1 & 4.6 & 0\\
		\hline
		\cite{divvala2014learning} & web & 36.2 & 10.3 & 10.3 & \textbf{12.5} & 27.5 & 1.5 & 6.5 & 10.0 & 6.0 & 16.4\\
		\hline
		\textbf{Ours} & \textbf{web} & 37.1 & \textbf{12.3} & \textbf{11.2} & 11.7 & 26.2 & \textbf{1.9} & 6.4 & \textbf{12.9} & \textbf{7.2} & \textbf{20.3}\\	
		\hline
		\cite{felzenszwalb2010object} & full & 59.3 & 15.7 & 35.9 & 10.3 & 48.5 & 33.2 & 26.9 & 20.2 & 43.3 & 16.4\\
		\hline				
	\end{tabular}
\end{table*}

\subsection{Model Learning}
Through the above steps, we firstly obtain the different ``visual patterns" for the given concept from the perspective of the text. Then for each ``visual pattern", we acquire relatively clean image dataset for the given concept by purifying noisy images. We put all the clean images together as training data for the given concept, and fine-tune a CNN model with a pre-trained model ``$bvlc\rule[-2pt]{0.15cm}{0.5pt}reference\rule[-2pt]{0.15cm}{0.5pt}caffenet$" \cite{jia2014caffe}.

\section{Experiments and Analysis}

Our proposed approach is a general framework that can be used to build a high-quality training set and train a robust recognition model for any given visual concept. To quantitatively evaluate the performance of our approach, we choose the \emph{Pascal VOC 2007} test set 20 categories for testing. Table 5 displays the results obtained using our algorithm and compares them with state-of-the-art baselines \cite{divvala2014learning,felzenszwalb2010object,siva2011weakly}.

\cite{siva2011weakly} is trained on \emph{Pascal VOC 2007} training data with image-level labels. \cite{divvala2014learning} is the state-of-the-art results for webly-supervised detection. \cite{felzenszwalb2010object} is the state-of-the-art results for fully-supervised detection. Compared to \cite{siva2011weakly} which uses weak supervision and \cite{felzenszwalb2010object} which uses full supervision, our method uses web-supervision as even the training set does not need to be labeled manually. Nonetheless, our results substantially surpass the previous best results in weakly supervised object detection.

Compared to \cite{divvala2014learning} which also uses web supervision, our method surpasses their results in most of the cases. The main reason for this is that our training data generated from the Internet contains much richer and accurate visual patterns in images. By observing the binding data in Table 4 and Table 5, we found that those concepts which have good performance tend to have sufficient word variations for the concept (with the exception of ``table''). In other words, our approach discovers concepts that have much more useful linkages to the visual patterns in the corresponding image set.

Lastly, we reach an important conclusion: a good training set should achieve successful results both in scale and in quality.

\section{Conclusion and Future Work}

We presented a fully automated approach to the generation of high-quality training data for any given visual concept. Our aim was to remove the need to laboriously annotate training datasets and train a robust recognition model with the generated training data for the given concept. Through our experiments on the benchmark Pascal VOC 2007 test set, we found our approach surpasses most of the previous best result in weakly supervised and webly supervised object detection. 
Using related word variations which link to the different ``visual patterns'' of images and then building the training image set for a concept or query according to these word variations is our first attempt to make use of textual metadata to build the training set. There is still room to improve our approach, for example, we can potentially use more sophisticated approaches to purify noisy images downloaded from the Internet and this will be the focus of our future work.


\begin{thebibliography}{1}\itemsep=-1pt\small

	\bibitem{chen2013neil}
	X.~Chen, A.~Shrivastava, and A.~Gupta.
	\newblock Neil: Extracting visual knowledge from web data.
	\newblock In {\em ICCV}, 2013.
	
	\bibitem{cilibrasi2007google}
	R.~L. Cilibrasi and P.~M. Vitanyi.
	\newblock The google similarity distance.
	\newblock In {\em TKDE}, 2007.
	
	\bibitem{collosal2005well}
	C.~Collosal.
	\newblock How well does the world wide web represent human language.
	\newblock In {\em Economist}, 2005.
	
	\bibitem{yao2018aaai}
	Y Yao, J Zhang, F Shen, W Yang, P Huang and Z Tang,
	\newblock Discovering and Distinguishing Multiple Visual Senses for Polysemous Words.
	\newblock In \textit{AAAI}, 2018.
	
	\bibitem{shen2018tmm}
	Y Yao, F Shen, J Zhang, L Liu, Z Tang and L Shao,
	\newblock Extracting Multiple Visual Senses for Web Learning
	\newblock In \textit{TMM}, 2019.
	
	\bibitem{shen2018tmm2}
	F Shen, J Zhang, L Liu, Z Tang and L Shao,
	\newblock Discovering and Distinguishing Multiple Visual Senses for Web Learning
	\newblock In \textit{TMM}, 2019.
	
	\bibitem{deng2009imagenet}
	J.~Deng, W.~Dong, R.~Socher, L.-J. Li, K.~Li, and L.~Fei-Fei.
	\newblock Imagenet: A large-scale hierarchical image database.
	\newblock In {\em CVPR}, 2009.
	
	\bibitem{divvala2014learning}
	S.~K. Divvala, A.~Farhadi, and C.~Guestrin.
	\newblock Learning everything about anything: Webly-supervised visual concept
	learning.
	\newblock In {\em CVPR}, 2014.
	
	\bibitem{doersch2012makes}
	C.~Doersch, S.~Singh, A.~Gupta, J.~Sivic, and A.~Efros.
	\newblock What makes paris look like paris?
	\newblock In {\em TOG}, 2012.
	
	\bibitem{fan2008liblinear}
	R.-E. Fan, K.-W. Chang, C.-J. Hsieh, X.-R. Wang, and C.-J. Lin.
	\newblock Liblinear: A library for large linear classification.
	\newblock In {\em JMLR}, 2008.
	
	\bibitem{xuicme2016}
	J Xu, J Zhang, F Shen, X Hua, and Z Tang,
	\newblock Automatic Image Dataset Construction with Multiple Textual Metadata.
	\newblock In \textit{ICME}, 2016.
	
	\bibitem{liutip2019}
	Y Yao, F Shen, J Zhang, L Liu, Z Tang and L Shao,
	\newblock Extracting Privileged Information for Enhancing Classifier Learning.
	\newblock In \textit{TIP}, 2019.
	
	\bibitem{tangijcai2018}
	Y Yao, J Zhang, F Shen, W Yang, X Hua and Z Tang,
	\newblock Extracting Privileged Information from Untagged Corpora for Classifier Learning.
	\newblock In \textit{IJCAI}, 2018.
	
	\bibitem{felzenszwalb2010object}
	P.~F. Felzenszwalb, R.~B. Girshick, D.~McAllester, and D.~Ramanan.
	\newblock Object detection with discriminatively trained part-based models.
	\newblock In {\em TPAMI}, 2010.
	
	\bibitem{fergus2010learning}
	R.~Fergus, L.~Fei-Fei, P.~Perona, and A.~Zisserman.
	\newblock Learning object categories from internet image searches.
	\newblock {\em Proceedings of the IEEE}, 2010.
	
	\bibitem{fergus2004visual}
	R.~Fergus, P.~Perona, and A.~Zisserman.
	\newblock A visual category filter for google images.
	\newblock In {\em ECCV}, 2004.
	
	\bibitem{hariharan2012discriminative}
	B.~Hariharan, J.~Malik, and D.~Ramanan.
	\newblock Discriminative decorrelation for clustering and classification.
	\newblock In {\em ECCV}, 2012.
	
	\bibitem{jia2014caffe}
	Y.~Jia, E.~Shelhamer, J.~Donahue, S.~Karayev, J.~Long, R.~Girshick,
	S.~Guadarrama, and T.~Darrell.
	\newblock Caffe: Convolutional architecture for fast feature embedding.
	\newblock In {\em MM}, 2014.
	
	\bibitem{kankanhalli1996cluster}
	M.~S. Kankanhalli, B.~M. Mehtre, and R.~K. Wu.
	\newblock Cluster-based color matching for image retrieval.
	\newblock In {\em PR}, 1996.
	
	\bibitem{li2010optimol}
	L.-J. Li and L.~Fei-Fei.
	\newblock Optimol: automatic online picture collection via incremental model
	learning.
	\newblock {\em IJCV}, 2010.
	
	\bibitem{lin2011large}
	Y.~Lin, F.~Lv, S.~Zhu, M.~Yang, T.~Cour, K.~Yu, L.~Cao, and T.~Huang.
	\newblock Large-scale image classification: fast feature extraction and svm
	training.
	\newblock In {\em CVPR}, 2011.
	
	\bibitem{lin2012syntactic}
	Y.~Lin, J.-B. Michel, E.~L. Aiden, J.~Orwant, W.~Brockman, and S.~Petrov.
	\newblock Syntactic annotations for the google books ngram corpus.
	\newblock In {\em ACL}, 2012.
	
	\bibitem{lucchi2012joint}
	A.~Lucchi and J.~Weston.
	\newblock Joint image and word sense discrimination for image retrieval.
	\newblock In {\em ECCV}, 2012.
	
	\bibitem{malisiewicz2008recognition}
	T.~Malisiewicz, A.~Efros, et~al.
	\newblock Recognition by association via learning per-exemplar distances.
	\newblock In {\em CVPR}, 2008.
	
	\bibitem{mezuman2012learning}
	E.~Mezuman and Y.~Weiss.
	\newblock Learning about canonical views from internet image collections.
	\newblock In {\em NIPS}, 2012.
	
	\bibitem{michel2011quantitative}
	J.-B. Michel, Y.~K. Shen, A.~P. Aiden, A.~Veres, M.~K. Gray, J.~P. Pickett,
	D.~Hoiberg, D.~Clancy, P.~Norvig, J.~Orwant, et~al.
	\newblock Quantitative analysis of culture using millions of digitized books.
	\newblock In {\em Science}, 2011.
	
	\bibitem{huatmm2017}
	Y Yao, J Zhang, F Shen, X Hua, J Xu and Z Tang,
	\newblock Exploiting Web Images for Dataset Construction: A Domain Robust Approach.
	\newblock In \textit{TMM}, 2017.
	
	\bibitem{zhang2016domain}
	Y Yao, X Hua, F Shen, J Zhang and Z Tang,
	\newblock A domain robust approach for image dataset construction.
	\newblock In \textit{ACM MM}, 2016. 
	
	\bibitem{prl2018}
	W Yang, P Huang, Q Wang, Y Cai and Z Tang,
	\newblock ``Exploiting Textual and Visual Features for Image Categorization''	
	\newblock In {\em PRL}, 2019.
	
	\bibitem{miller1995wordnet}
	G.~A. Miller.
	\newblock Wordnet: a lexical database for english.
	\newblock In {\em Communications of the ACM}, 1995.
	
	\bibitem{perona2010vision}
	P.~Perona.
	\newblock Vision of a visipedia.
	\newblock In {\em Proceedings of the IEEE}, 2010.
	
	\bibitem{prest2012learning}
	A.~Prest, C.~Leistner, J.~Civera, C.~Schmid, and V.~Ferrari.
	\newblock Learning object class detectors from weakly annotated video.
	\newblock In {\em CVPR}, 2012.
	
	\bibitem{rubinstein2013unsupervised}
	M.~Rubinstein, A.~Joulin, J.~Kopf, and C.~Liu.
	\newblock Unsupervised joint object discovery and segmentation in internet
	images.
	\newblock In {\em CVPR}, 2013.
	
	\bibitem{schroff2011harvesting}
	F.~Schroff, A.~Criminisi, and A.~Zisserman.
	\newblock Harvesting image databases from the web.
	\newblock In {\em TPAMI}, 2011.
	
	\bibitem{Fumin_2015_ICCV}
	F.~Shen, W.~Liu, S.~Zhang, Y.~Yang, and H.~Shen.
	\newblock Learning binary codes for maximum inner product search.
	\newblock In {\em ICCV}, 2015.
	
	\bibitem{Shen_2015_CVPR}
	F.~Shen, C.~Shen, W.~Liu, and H.~Tao~Shen.
	\newblock Supervised discrete hashing.
	\newblock In {\em CVPR}, 2015.
	
	\bibitem{shen2013inductive}
	F.~Shen, C.~Shen, Q.~Shi, A.~Van Den~Hengel, and Z.~Tang.
	\newblock Inductive hashing on manifolds.
	\newblock In {\em CVPR}, 2013.
	
	\bibitem{siddiquie2010beyond}
	B.~Siddiquie and A.~Gupta.
	\newblock Beyond active noun tagging: Modeling contextual interactions for
	multi-class active learning.
	\newblock In {\em CVPR}, 2010.
	
	\bibitem{siva2011weakly}
	P.~Siva and T.~Xiang.
	\newblock Weakly supervised object detector learning with model drift
	detection.
	\newblock In {\em ICCV}, 2011.
	
	\bibitem{vijayanarasimhan2014large}
	S.~Vijayanarasimhan and K.~Grauman.
	\newblock Large-scale live active learning: Training object detectors with
	crawled data and crowds.
	\newblock {\em IJCV}, 2014.
	
	\bibitem{volkel2006semantic}
	M.~V{\"o}lkel, M.~Kr{\"o}tzsch, D.~Vrandecic, H.~Haller, and R.~Studer.
	\newblock Semantic wikipedia.
	\newblock In {\em WWW}, 2006.
	
	\bibitem{wang2007cell}
	W.~Wang and H.~Song.
	\newblock Cell cluster image segmentation on form analysis.
	\newblock In {\em ACL}, 2007.
	
	\bibitem{you2015robust}
	Q.~You, J.~Luo, H.~Jin, and J.~Yang.
	\newblock Robust image sentiment analysis using progressively trained and
	domain transferred deep networks.
	\newblock In {\em AAAI}, 2015.
	
	
	

	

	

	
	\bibitem{mta2018}
	H Liu, X Han, X Li, Y Yao, P Huang, Z Tang,
	\newblock ``Deep Representation Learning for Road Detection using Siamese Network''	
	\newblock In {\em MTA}, 2018.
	
	\bibitem{huangpu}
	P Huang, T Li, G Gao, Y Yao and G Yang,
	\newblock ``Collaborative Representation Based Local Discriminant Projection for Feature Extraction''	
	\newblock In {\em DSP}, 2018.
	
	\bibitem{yaotkde2019}
	Y Yao, J Zhang, F Shen, L Liu, F Zhu, and H Shen,
	\newblock Towards Automatic Construction of Diverse, High-quality Image Datasets.
	\newblock In \textit{TKDE}, 2019. 
	
	\bibitem{xu2017}
	M Xu, Z Tang, Y Yao, L Yao, H Liu and J Xu,
	\newblock ``Deep Learning for Person Reidentification Using Support Vector Machines''	
	\newblock In {\em AIM}, 2017.
	
	\bibitem{tang2017}
	Y Yao, J Zhang, F Shen, X Hua, J Xu and Z Tang,
	\newblock ``A New Web-supervised Method for Image Dataset Constructions''	
	\newblock In {\em NEUROCOM}, 2017.
	
\end{thebibliography}
\end{document}